\documentstyle[aclap,psfig,psfig]{article}

\newtheorem{example1}{Example}
\newenvironment{example}[1]{\begin{example1}[#1] \small\mbox{} \\ \rm}{\end{example1}}

\begin{document}

\title{Intonational Boundaries, Speech Repairs and \\
Discourse Markers: Modeling Spoken Dialog\thanks{In Proceedings of ACL/EACL'97}}

\author{Peter A. Heeman and James F. Allen \\
Department of Computer Science \\
University of Rochester \\
Rochester NY 14627, USA \\
{\tt \{heeman,james\}@cs.rochester.edu}}

\maketitle

\begin{abstract}
%
To understand a speaker's turn of a conversation, one needs to segment
it into intonational phrases, clean up any speech repairs that might
have occurred, and identify discourse markers. In this paper, we argue
that these problems must be resolved together, and that they must be
resolved early in the processing stream. We put forward a statistical
language model that resolves these problems, does POS tagging, and can
be used as the language model of a speech recognizer. We find that by
accounting for the interactions between these tasks that the
performance on each task improves, as does POS tagging and perplexity.
\end{abstract}

\newcommand{\comment}[1]{}

\newcommand{\mc}[1]{\multicolumn{1}{c|}{#1}}

\newcommand{\interruptionpoint}{
\parbox[t]{0.7em}{
\raisebox{-1.4em}{\hspace{-.4em}\LARGE\bf$\uparrow$} \vspace{-0.2em} \\
\makebox[1em][c]{\small\em interruption point} 
}}

\newcommand{\ip}{
\parbox[t]{0.5em}{
\raisebox{-1.4em}{\hspace{-.4em}\LARGE\bf$\uparrow$} \vspace{-0.4em} \\
\hspace*{-0.4em}{\small\em ip}}}

\newcommand{\reparandum}[1]{$\underbrace{\makebox{#1}}_{\makebox{\small\em reparandum}}$}
\newcommand{\editingterm}[1]{$\underbrace{\makebox{#1}}_{\makebox{\small\em
editing term}}$}
\newcommand{\et}[1]{$\underbrace{\makebox{#1}}_{\makebox{\small\em
et}}$}
\newcommand{\alteration}[1]{$\underbrace{\makebox{#1}}_{\makebox{\small\em alteration}}$}

\newcommand{\etal}{{\em et al.}}
\newcommand{\dash}{\mbox{-}}
\newcommand{\rim}{1\!,i\mbox{-}1}
\newcommand{\ri}{1\!,i}

\section{Introduction}

Interactive spoken dialog provides many new challenges for natural
language understanding systems.  One of the most critical challenges
is simply determining the speaker's intended utterances: both
segmenting the speaker's turn into utterances and determining the
intended words in each utterance.  Since there is no well-agreed to
definition of what an utterance is, we instead focus on intonational
phrases \cite {silverman-etal:92:icslp}, which end with an
acoustically signaled {\em boundary tone}.  Even assuming perfect word
recognition, the problem of determining the intended words is
complicated due to the occurrence of speech repairs, which occur where
the speaker goes back and changes (or repeats) something she just
said.  The words that are replaced or repeated are no longer part of
the intended utterance, and so need to be identified.  The following
example, from the Trains corpus \cite{HeemanAllen95:cdrom}, gives an
example of a speech repair with the words that the speaker intends to
be replaced marked by {\em reparandum}, the words that are the
intended replacement marked as {\em alteration}, and the cue phrases
and filled pauses that tend to occur in between marked as the {\em
editing term}.
\begin{example}{d92a-5.2 utt34} 
we'll pick up 
$\underbrace{\makebox{a tank of}}_{\makebox[2.5em][r]{\footnotesize\em reparandum \hfill}}$\interruptionpoint$\underbrace{\makebox{uh}}_{\hfill\makebox[4em][l]{\footnotesize\em editing term}}$
$\underbrace{\makebox{the tanker of}}_{\makebox[2.5em][l]{\hfill\footnotesize\em
alteration}}$ \makebox[1em][l]{oranges}
\end{example}

Much work has been done on both detecting boundary tones
(e.g.~\cite{WangHirschberg92:csl,WightmanOstendorf94:ieee,StolckeShriberg96:icslp,Kompe-etal94:icassp,Mast-etal96:icslp})
and on speech repair detection and correction
(e.g.~\cite{Hindle83:acl,Bear-etal92:acl,NakataniHirschberg94:jasa,HeemanAllen94:acl,StolckeShriberg96:icassp}).
This work has focused on one of the issues in isolation of the other.
However, these two issues are intertwined.  Cues such as the presence
of silence, final syllable lengthening, and presence of filled pauses
tend to mark both events.
Even the presence of word correspondences, a tradition cue for
detecting and correcting speech repairs, sometimes marks boundary
tones as well, as illustrated by the following example where the
intonational phrase boundary is marked with the ToBI symbol {\bf \%}.
\begin{example}{d93-83.3~utt73}
that's all you need {\bf \%} you only need one boxcar
\end{example} 

Intonational phrases and speech repairs also interact with the
identification of discourse markers. Discourse markers
\cite{Schiffrin87:book,HirschbergLitman93:cl,ByronHeeman97:eurospeech}
are used to relate new speech to the current discourse state.  Lexical
items that can function as discourse markers, such as ``well'' and
``okay,'' are ambiguous as to whether they are being used as discourse
markers or not.  The complication is that discourse markers tend to be
used to introduce a new utterance, or can be an utterance all to
themselves (such as the acknowledgment ``okay'' or ``alright''), or
can be used as part of the editing term of a speech repair, or to
begin the alteration.  Hence, the problem of identifying discourse
markers also needs to be addressed with the segmentation and speech
repair problems.

These three phenomena of spoken dialog, however, cannot be resolved
without recourse to syntactic information.  Speech repairs, for
example, are often signaled by syntactic anomalies.  Furthermore, in
order to determine the extent of the reparandum, one needs to take
into account the parallel structure that typically exists between the
reparandum and alteration, which relies on at identifying the
syntactic roles, or part-of-speech (POS) tags, of the words involved
\cite{Bear-etal92:acl,HeemanAllen94:acl}.  However, speech repairs
disrupt the context that is needed to determine the POS tags
\cite{Hindle83:acl}.  Hence, speech repairs, as well as boundary tones
and discourse markers, must be resolved during syntactic
disambiguation.

Of course when dealing with spoken dialogue, one cannot forget the
initial problem of determining the actual words that the speaker is
saying. Speech recognizers rely on being able to predict the
probability of what word will be said next. Just as intonational
phrases and speech repairs disrupt the local context that is needed
for syntactic disambiguation, the same holds for predicting what word
will come next. If a speech repair or intonational phrase occurs, this
will alter the probability estimate. But more importantly, speech
repairs and intonational phrases have acoustic correlates such as the
presence of silence. Current speech recognition language models cannot
account for the presence of silence, and tend to simply ignore it. By
modeling speech repairs and intonational boundaries, we can take into
account the acoustic correlates and hence use more of the available
information.

From the above discussion, it is clear that we need to model these
dialogue phenomena together and very early on in the speech processing
stream, in fact, during speech recognition.  Currently, the approaches
that work best in speech recognition are statistical approaches that
are able to assign probability estimates for what word will occur next
given the previous words.  Hence, in this paper, we introduce a
statistical language model that can detect speech repairs, boundary
tones, and discourse markers, and can assign POS tags, and can use
this information to better predict what word will occur next.

In the rest of the paper, we first introduce the Trains corpus.  We
then introduce a statistical language model that incorporates POS
tagging and the identification of discourse markers.  We then augment
this model with speech repair detection and correction and
intonational boundary tone detection.  We then present the results of
this model on the Trains corpus and show that it can better account
for these discourse events than can be achieved by modeling them
individually.  We also show that by modeling these two phenomena that
we can increase our POS tagging performance by 8.6\%, and improve our
ability to predict the next word.

\section{Trains Corpus}

As part of the {\sc Trains} project \cite{Allen-etal95:jetai-s}, which
is a long term research project to build a conversationally proficient
planning assistant, we have collected a corpus of problem solving
dialogs \cite{HeemanAllen95:cdrom}.  The dialogs involve two
human participants, one who is playing the role of a user and has a
certain task to accomplish, and another who is playing the role of the
system by acting as a planning assistant.  The collection methodology
was designed to make the setting as close to human-computer
interaction as possible, but was not a {\em wizard} scenario, where
one person pretends to be a computer.  Rather, the user knows that he
is talking to another person.

The {\sc Trains} corpus consists of about six and half hours of
speech. Table~\ref{tab:occurrences} gives some general statistics
about the corpus, including the number of dialogs, speakers, words,
speaker turns, and occurrences of discourse markers, boundary tones
and speech repairs.
\begin{table}
\begin{center}
{\small\begin{tabular}{|l|r|}\hline
Dialogs                      &    98 \\
Speakers                     &    34 \\
Words                        & 58298 \\
Turns                        &  6163 \\
Discourse Markers            &  8278 \\
Boundary Tones               & 10947 \\
Turn-Internal Boundary Tones &  5535 \\
Abridged Repairs             &   423 \\
Modification Repairs         &  1302 \\
Fresh Starts                 &   671 \\
Editing Terms                &  1128 \\ \hline
\end{tabular}}
\end{center}
\caption{Frequency of Tones, Repairs and Editing Terms in the Trains Corpus}
\label{tab:occurrences}
\end{table}

The speech repairs in the Trains corpus have been hand-annotated.
We have divided the repairs into three types: {\em fresh starts}, {\em
modification repairs}, and {\em abridged repairs}.\footnote{This
classification is similar to that of Hindle \shortcite{Hindle83:acl} and
Levelt \shortcite{Levelt83:cog}.}  A fresh start is where the speaker
abandons the current utterance and starts again, where the abandonment
seems acoustically signaled.
\begin{example}{d93-12.1 utt30}
\reparandum{so it'll take}\interruptionpoint \editingterm{um}
\alteration{so you want to do what}
\end{example}
The second type of repairs are the modification repairs.  These
include all other repairs in which the reparandum is not empty.
\begin{example}{d92a-1.3 utt65}
so that \reparandum{will total}\interruptionpoint
\alteration{will take} seven hours to do that
\end{example}
The third type of repairs are the abridged repairs, which consist
solely of an editing term. Note that utterance initial filled pauses
are not treated as abridged repairs.
\begin{example}{d93-14.3 utt42}
we need to\interruptionpoint$\!\!\underbrace{\makebox{um}}_{\hfill\makebox[1.5em][l]{\small\em editing term}}\!$ manage to get the bananas to Dansville
\end{example}

There is typically a correspondence between the reparandum and the
alteration, and following Bear~\etal~\shortcite {Bear-etal92:acl}, we
annotate this using the labels {\bf m} for word matching and {\bf r}
for word replacements (words of the same syntactic category). Each
pair is given a unique index. Other words in the reparandum and
alteration are annotated with an {\bf x}. Also, editing terms (filled
pauses and clue words) are labeled with {\bf et}, and the interruption
point with {\bf ip}, which will occur before any editing terms
associated with the repair, and after a word fragment, if present. The
interruption point is also marked as to whether the repair is a
fresh start, modification repair, or abridged repair, in which cases,
we use {\bf ip:can}, {\bf ip:mod} and {\bf ip:abr}, respectively. The
example below illustrates how a repair is annotated in this scheme.
\begin{example}{d93-15.2 utt42} \vspace{-0.4em}
\label{ex:d93-15.2:utt42}
\setlength{\tabcolsep}{0.15em}
\begin{tabular}{cccccccccc}
engine & two & from & Elmi(ra)- & & or & engine  & three & from & Elmira \\
\small\bf m1 & \small\bf r2 & \small\bf m3 & \small\bf m4 & \makebox[0.1em][c]{\Large\bf$\uparrow$} & \small\bf
et & \small\bf m1 & \small\bf r2 & \small\bf m3 & \small\bf m4 \\
       &     & & & \makebox[0.1em][r]{\small\bf ip:mod}   &   &    &    &         \\
\end{tabular} 
\end{example}

\section{A POS-Based Language Model}

The goal of a speech recognizer is to find the sequence of words
$\hat{W}$ that is maximal given the acoustic signal $A$. However, for
detecting and correcting speech repairs, and identifying boundary
tones and discourse markers, we need to augment the model so that it
incorporates shallow statistical analysis, in the form of POS tagging.
The POS tagset, based on the Penn Treebank tagset \cite
{Marcus-etal93:cl}, includes special tags for denoting when a word is
being used as a discourse marker. In this section, we give an overview
of our basic language model that incorporates POS tagging. Full
details can be found in \cite
{HeemanAllen97:eurospeech,Heeman97:thesis}.

To add in POS tagging, we change the goal of the speech recognition
process to find the best word and POS tags given the acoustic signal.
The derivation of the acoustic model and language model is now as
follows.
\begin{eqnarray*}
\hat{W}\hat{P}&=&\arg\max_{W,P}\Pr(WP|A)\\
&=&\arg\max_{WP}\frac{\Pr(A|WP)\Pr(WP)}{\Pr(A)}\\ 
&=&\arg\max_{WP}\Pr(A|WP)\Pr(WP)
\end{eqnarray*}
The first term $\Pr(A|WP)$ is the factor due to the acoustic model,
which we can approximate by $\Pr(A|W)$. The second term $\Pr(WP)$ is
the factor due to the language model. We rewrite $\Pr(WP)$ as
$\Pr(W_{1,N}P_{1,N})$, where $N$ is the number of words in the
sequence. We now rewrite the language model probability as follows.
\begin{eqnarray*}
\lefteqn{\Pr(W_{1,N}P_{1,N})} \\
&=&\prod_{i=1,N}\Pr(W_iP_i|W_{\rim}P_{\rim})\\
&=&\prod_{i=1,N}\Pr(W_i|W_{\rim}P_{\ri})\Pr(P_i|W_{\rim}P_{\rim})
\end{eqnarray*}
We now have two probability distributions that we need to estimate,
which we do using decision trees \cite
{Breiman-etal84:book,Bahl-etal89:tassp}. The decision tree algorithm
has the advantage that it uses information theoretic measures to
construct equivalence classes of the context in order to cope with
sparseness of data. The decision tree algorithm starts with all of the
training data in a single leaf node. For each leaf node, it looks for
the question to ask of the context such that splitting the node
into two leaf nodes results in the biggest decrease in {\em impurity},
where the impurity measures how well each leaf predicts the events in
the node. After the tree is grown, a heldout dataset is used to smooth
the probabilities of each node with its parent
\cite{Bahl-etal89:tassp}.

\begin{figure}
\begin{center}
\begin{picture}(0,0)%
\includegraphics{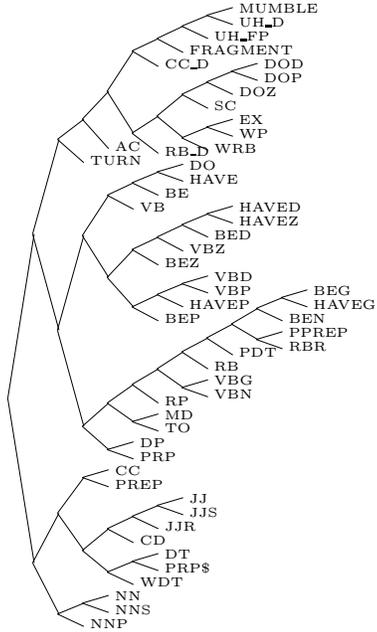}%
\end{picture}%
\setlength{\unitlength}{0.008125in}%
\begingroup\makeatletter\ifx\SetFigFont\undefined
\def\x#1#2#3#4#5#6#7\relax{\def\x{#1#2#3#4#5#6}}%
\expandafter\x\fmtname xxxxxx\relax \def\y{splain}%
\ifx\x\y   
\gdef\SetFigFont#1#2#3{%
  \ifnum #1<17\tiny\else \ifnum #1<20\small\else
  \ifnum #1<24\normalsize\else \ifnum #1<29\large\else
  \ifnum #1<34\Large\else \ifnum #1<41\LARGE\else
     \huge\fi\fi\fi\fi\fi\fi
  \csname #3\endcsname}%
\else
\gdef\SetFigFont#1#2#3{\begingroup
  \count@#1\relax \ifnum 25<\count@\count@25\fi
  \def\x{\endgroup\@setsize\SetFigFont{#2pt}}%
  \expandafter\x
    \csname \romannumeral\the\count@ pt\expandafter\endcsname
    \csname @\romannumeral\the\count@ pt\endcsname
  \csname #3\endcsname}%
\fi
\fi\endgroup
\begin{picture}(193,413)(1,422)
\put(146,819){\makebox(0,0)[lb]{\smash{\SetFigFont{5}{6.0}{rm} MUMBLE}}}
\put(146,810){\makebox(0,0)[lb]{\smash{\SetFigFont{5}{6.0}{rm} UH\_D}}}
\put(130,801){\makebox(0,0)[lb]{\smash{\SetFigFont{5}{6.0}{rm} UH\_FP}}}
\put(114,792){\makebox(0,0)[lb]{\smash{\SetFigFont{5}{6.0}{rm} FRAGMENT}}}
\put( 98,783){\makebox(0,0)[lb]{\smash{\SetFigFont{5}{6.0}{rm} CC\_D}}}
\put(162,783){\makebox(0,0)[lb]{\smash{\SetFigFont{5}{6.0}{rm} DOD}}}
\put(162,774){\makebox(0,0)[lb]{\smash{\SetFigFont{5}{6.0}{rm} DOP}}}
\put(146,765){\makebox(0,0)[lb]{\smash{\SetFigFont{5}{6.0}{rm} DOZ}}}
\put(130,756){\makebox(0,0)[lb]{\smash{\SetFigFont{5}{6.0}{rm} SC}}}
\put(146,747){\makebox(0,0)[lb]{\smash{\SetFigFont{5}{6.0}{rm} EX}}}
\put(146,738){\makebox(0,0)[lb]{\smash{\SetFigFont{5}{6.0}{rm} WP}}}
\put(130,729){\makebox(0,0)[lb]{\smash{\SetFigFont{5}{6.0}{rm} WRB}}}
\put( 98,727){\makebox(0,0)[lb]{\smash{\SetFigFont{5}{6.0}{rm} RB\_D}}}
\put( 66,730){\makebox(0,0)[lb]{\smash{\SetFigFont{5}{6.0}{rm} AC}}}
\put( 50,721){\makebox(0,0)[lb]{\smash{\SetFigFont{5}{6.0}{rm} TURN}}}
\put(114,718){\makebox(0,0)[lb]{\smash{\SetFigFont{5}{6.0}{rm} DO}}}
\put(114,709){\makebox(0,0)[lb]{\smash{\SetFigFont{5}{6.0}{rm} HAVE}}}
\put( 98,700){\makebox(0,0)[lb]{\smash{\SetFigFont{5}{6.0}{rm} BE}}}
\put( 82,691){\makebox(0,0)[lb]{\smash{\SetFigFont{5}{6.0}{rm} VB}}}
\put(146,691){\makebox(0,0)[lb]{\smash{\SetFigFont{5}{6.0}{rm} HAVED}}}
\put(146,682){\makebox(0,0)[lb]{\smash{\SetFigFont{5}{6.0}{rm} HAVEZ}}}
\put(130,673){\makebox(0,0)[lb]{\smash{\SetFigFont{5}{6.0}{rm} BED}}}
\put(114,664){\makebox(0,0)[lb]{\smash{\SetFigFont{5}{6.0}{rm} VBZ}}}
\put( 98,655){\makebox(0,0)[lb]{\smash{\SetFigFont{5}{6.0}{rm} BEZ}}}
\put(130,646){\makebox(0,0)[lb]{\smash{\SetFigFont{5}{6.0}{rm} VBD}}}
\put(130,637){\makebox(0,0)[lb]{\smash{\SetFigFont{5}{6.0}{rm} VBP}}}
\put(114,628){\makebox(0,0)[lb]{\smash{\SetFigFont{5}{6.0}{rm} HAVEP}}}
\put( 98,619){\makebox(0,0)[lb]{\smash{\SetFigFont{5}{6.0}{rm} BEP}}}
\put(194,637){\makebox(0,0)[lb]{\smash{\SetFigFont{5}{6.0}{rm} BEG}}}
\put(194,628){\makebox(0,0)[lb]{\smash{\SetFigFont{5}{6.0}{rm} HAVEG}}}
\put(178,619){\makebox(0,0)[lb]{\smash{\SetFigFont{5}{6.0}{rm} BEN}}}
\put(178,610){\makebox(0,0)[lb]{\smash{\SetFigFont{5}{6.0}{rm} PPREP}}}
\put(178,601){\makebox(0,0)[lb]{\smash{\SetFigFont{5}{6.0}{rm} RBR}}}
\put(146,597){\makebox(0,0)[lb]{\smash{\SetFigFont{5}{6.0}{rm} PDT}}}
\put(130,588){\makebox(0,0)[lb]{\smash{\SetFigFont{5}{6.0}{rm} RB}}}
\put(130,579){\makebox(0,0)[lb]{\smash{\SetFigFont{5}{6.0}{rm} VBG}}}
\put(130,570){\makebox(0,0)[lb]{\smash{\SetFigFont{5}{6.0}{rm} VBN}}}
\put( 98,566){\makebox(0,0)[lb]{\smash{\SetFigFont{5}{6.0}{rm} RP}}}
\put( 98,557){\makebox(0,0)[lb]{\smash{\SetFigFont{5}{6.0}{rm} MD}}}
\put( 98,548){\makebox(0,0)[lb]{\smash{\SetFigFont{5}{6.0}{rm} TO}}}
\put( 82,539){\makebox(0,0)[lb]{\smash{\SetFigFont{5}{6.0}{rm} DP}}}
\put( 82,530){\makebox(0,0)[lb]{\smash{\SetFigFont{5}{6.0}{rm} PRP}}}
\put( 66,521){\makebox(0,0)[lb]{\smash{\SetFigFont{5}{6.0}{rm} CC}}}
\put( 66,512){\makebox(0,0)[lb]{\smash{\SetFigFont{5}{6.0}{rm} PREP}}}
\put(114,503){\makebox(0,0)[lb]{\smash{\SetFigFont{5}{6.0}{rm} JJ}}}
\put(114,494){\makebox(0,0)[lb]{\smash{\SetFigFont{5}{6.0}{rm} JJS}}}
\put( 98,485){\makebox(0,0)[lb]{\smash{\SetFigFont{5}{6.0}{rm} JJR}}}
\put( 82,476){\makebox(0,0)[lb]{\smash{\SetFigFont{5}{6.0}{rm} CD}}}
\put( 98,467){\makebox(0,0)[lb]{\smash{\SetFigFont{5}{6.0}{rm} DT}}}
\put( 98,458){\makebox(0,0)[lb]{\smash{\SetFigFont{5}{6.0}{rm} PRP\$}}}
\put( 82,449){\makebox(0,0)[lb]{\smash{\SetFigFont{5}{6.0}{rm} WDT}}}
\put( 66,440){\makebox(0,0)[lb]{\smash{\SetFigFont{5}{6.0}{rm} NN}}}
\put( 66,431){\makebox(0,0)[lb]{\smash{\SetFigFont{5}{6.0}{rm} NNS}}}
\put( 50,422){\makebox(0,0)[lb]{\smash{\SetFigFont{5}{6.0}{rm} NNP}}}
\end{picture}

\end{center}
\caption{POS Classification Tree}
\label{fig:postags}
\end{figure}

To allow the decision tree to ask about the words and POS tags in the
context, we cluster the words and POS tags using the algorithm of
Brown~\etal~\shortcite {Brown-etal92:cl} into a binary classification
tree.  This gives an implicit binary encoding for each word and POS
tag, thus allowing the decision tree to ask about the words and POS
tags using simple binary questions, such as `is the third bit of the
POS tag encoding equal to one?'  Figure~\ref{fig:postags} shows a POS
classification tree.  The binary encoding for a POS tag is determined
by the sequence of top and bottom edges that leads from the root node
to the node for the POS tag.
%

Unlike other work (e.g.~\cite{Black-etal92:darpa:pos,Magerman95:acl}),
we treat the word identities as a further refinement of the POS tags;
thus we build a word classification tree for each POS tag.  This has
the advantage of avoiding unnecessary data fragmentation, since the
POS tags and word identities are no longer separate sources of
information.  As well, it constrains the task of building the word
classification trees since the major distinctions are captured by the
POS classification tree.

\section{Augmenting the Model}

Just as we redefined the speech recognition problem so as to account
for POS tagging and identifying discourse markers, we do the same for
modeling boundary tones and speech repairs. We introduce null tokens
between each pair of consecutive words $w_{i{\dash}1}$ and $w_i$
\cite{HeemanAllen94:acl}, which will be tagged as to the occurrence of
these events. The boundary tone tag $T_i$ indicates if word
$w_{i{\dash}1}$ ends an intonational boundary ($T_i$={\bf T}), or not
($T_i$={\bf null}).

For detecting speech repairs, we have the problem that repairs are
often accompanied by an editing term, such as ``um'', ``uh'',
``okay'', or ``well'', and these must be identified as such.
Furthermore, an editing term might be composed of a number of words,
such as ``let's see'' or ``uh well''. Hence we use two tags: an
editing term tag $E_i$ and a repair tag $R_i$. The editing term tag
indicates if $w_i$ starts an editing term ($E_i$={\bf Push}), if $w_i$
continues an editing term ($E_i$={\bf ET}), if $w_{i{\dash}1}$ ends an
editing term ($E_i$={\bf Pop}), or otherwise ($E_i$={\bf null}). The
repair tag $R_i$ indicates whether word $w_i$ is the onset of the
alteration of a fresh start ($R_i$={\bf C}), a modification repair
($R_i$={\bf M}), or an abridged repair ($R_i$={\bf A}), or there is
not a repair ($R_i$={\bf null}). Note that for repairs with an editing
term, the repair is tagged after the extent of the editing term has
been determined. Below we give an example showing all non-null tone,
editing term and repair tags.
\begin{example}{d93-18.1 utt47}
\label{ex:d93-18.1:utt47} it takes one {\bf Push} you {\bf ET} know
{\bf Pop} {\bf M} two hours {\bf T} 
\end{example}

If a modification repair or fresh start occurs, we need to determine
the extent (or the onset) of the reparandum, which we refer to as {\em
correcting} the speech repair. Often, speech repairs have strong word
correspondences between the reparandum and alteration, involving word
matches and word replacements. Hence, knowing the extent of the
reparandum means that we can use the reparandum to predict the words
(and their POS tags) that make up the alteration. For $R_i \in \{{\bf
Mod},{\bf Can}\}$, we define $O_i$ to indicate the onset of the
reparandum.\footnote {Rather than estimate $O_i$ directly, we instead
query each potential onset to see how likely it is to be the actual
onset of the reparandum.}

\begin{figure}
\centerline{\psfig{file=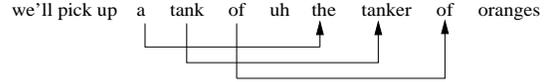,width=2.8in}} \caption{Cross Serial
Correspondences} \label{fig:cross} \end{figure}

If we are in the midst of processing a repair, we need to determine if
there is a word correspondence from the reparandum to the current word
$w_i$.  The tag $L_i$ is used to indicate which word in the reparandum
is licensing the correspondence.  Word correspondences tend to exhibit
a cross serial dependency; in other words if we have a correspondence
between $w_j$ in the reparandum and $w_k$ in the alteration, any
correspondence with a word in the alteration after $w_k$ will be to a
word that is after $w_j$, as illustrated in Figure~\ref{fig:cross}.
This means that if $w_i$ involves a word correspondence, it will most
likely be with a word that follows the last word in the reparandum
that has a word correspondence. Hence, we restrict $L_i$ to only those
words that are after the last word in the reparandum that has a
correspondence (or from the reparandum onset if there is not yet a
correspondence). If there is no word correspondence for $w_i$, we set
$L_i$ to the first word after the last correspondence.
	
The second tag involved in the correspondences is $C_i$, which
indicates the type of correspondence between the word indicated by
$L_i$ and the current word $w_i$.  We focus on word correspondences
that involve either a word match ($C_i$={\bf m}), a word replacement
($C_i$={\bf r}), where both words are of the same POS tag, or no
correspondence ($C_i$={\bf x}).

Now that we have defined these six additional tags for modeling
boundary tones and speech repairs, we redefine the speech recognition
problem so that its goal is to find the maximal assignment for the
words as well as the POS, boundary tone, and speech repair tags.
\newlength{\maxarg}
\settowidth{\maxarg}{$\arg\max$}
{\small
\[
\hat{W}\hat{P}\hat{C}\hat{L}\hat{O}\hat{R}\hat{E}\hat{T} = 
\parbox[t]{\maxarg}{$\arg\max$ \tiny \\ 
\makebox[5.5em][r]{\tiny $WPCLORET$}} \Pr(WCLORET|A)
\]}%
The result is that we now have eight probability distributions that we
need to estimate.  
{\small
\[\Pr(T_i|W_{\rim}P_{\rim}C_{\rim}L_{\rim}O_{\rim}R_{\rim}E_{\rim}T_{\rim}) \]
\[\Pr(E_i|W_{\rim}P_{\rim}C_{\rim}L_{\rim}O_{\rim}R_{\rim}E_{\rim}T_{\ri}) \]
\[\Pr(R_i|W_{\rim}P_{\rim}C_{\rim}L_{\rim}O_{\rim}R_{\rim}E_{\ri}T_{\ri}) \]
\[\Pr(O_i|W_{\rim}P_{\rim}C_{\rim}L_{\rim}O_{\rim}R_{\ri}E_{\ri}T_{\ri}) \]
\[\Pr(L_i|W_{\rim}P_{\rim}C_{\rim}L_{\rim}O_{\ri}R_{\ri}E_{\ri}T_{\ri}) \]
\[\Pr(C_i|W_{\rim}P_{\rim}C_{\rim}L_{\ri}O_{\ri}R_{\ri}E_{\ri}T_{\ri}) \]
\[\Pr(P_i|W_{\rim}P_{\rim}C_{\ri}L_{\ri}O_{\ri}R_{\ri}E_{\ri}T_{\ri}) \]
\[\Pr(W_i|W_{\rim}P_{\ri}C_{\ri}L_{\ri}O_{\ri}R_{\ri}E_{\ri}T_{\ri}) \]}

The context for each of the probability distributions includes all of
the previous context. In principal, we could give all of this context
to the decision tree algorithm and let it decide what information is
relevant in constructing equivalence classes of the contexts. However,
the amount of training data is limited (as are the learning
techniques) and so we need to encode the context in order to simplify
the task of constructing meaningful equivalence classes. We start with
the words and their POS tags that are in the context and for each
non-null tone, editing term (we also skip over $E$={\bf ET}), and
repair tag, we insert it into the appropriate place, just as
Kompe~\etal~\shortcite{Kompe-etal94:icassp} do for boundary tones in
their language model. Below we give the encoded context for the word
``know'' from Example~\ref{ex:d93-18.1:utt47}
\begin{example}{d93-18.1 utt47}
it/{\bf PRP} takes/{\bf VBP} one/{\bf CD} {\bf Push} you/{\bf PRP}
\end{example}
The result of this is that the non-null tag values are treated just as
if they were lexical items.\footnote{Since we treat the non-null tags
as lexical items, we associate a unique POS tag with each value.}
Furthermore, if an editing term is completed, or the extent of a
repair is known, we can also clean up the editing term or reparandum,
respectively, in the same way that Stolcke and Shriberg
\shortcite {StolckeShriberg96:icassp} clean up filled pauses, and
simple repair patterns. This means that we can then generalize between
fluent speech and instances that have a repair. For instance, in the
two examples below, the context for the word ``get'' and its POS tag
will be the same for both, namely ``so/{\bf CC\_D} we/{\bf PRP} need/{\bf VBP} to/{\bf TO}''.
\begin{example}{d93-11.1 utt46}
so we need to get the three tankers
\end{example}
\begin{example}{d92a-2.2 utt6}
so we need to {\bf Push} um {\bf Pop} {\bf A} get a tanker of OJ
\end{example}

We also include other features of the context. For instance, we
include a variable to indicate if we are currently processing an
editing term, and whether a non-filled pause editing term was seen.
For estimating $R_i$, we include the editing terms as well. For
estimating $O_i$, we include whether the proposed reparandum includes
discourse markers, filled pauses that are not part of an editing term,
boundary terms, and whether the proposed reparandum overlaps with any
previous repair.

\section{Silences}

\begin{figure}
\centerline{\psfig{figure=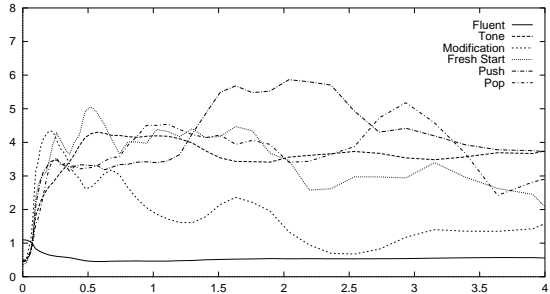,width=\columnwidth}}
\caption{Preference for tone, editing term, and repair tags given the length of silence}
\label{fig:silences}
\end{figure}

Silence, as well as other acoustic information, can also give evidence
as to whether an intonational phrase, speech repair, or editing term
occurred. We include $S_i$, the silence duration between word
$w_{i{\dash}1}$ and $w_i$, as part of the context for conditioning the
probability distributions for the tone $T_i$, editing term $E_i$, and
repair $R_i$ tags. Due to sparseness of data, we make several the
independence assumptions so that we can separate the silence
information from the rest of the context. For example, for the tone
tag, let ${\em Rest}_i$ represent the rest of the context that is
used to condition $T_i$. By assuming that ${\em Rest}_i$ and $S_i$
are independent, and are independent given $T_i$, we can rewrite
$\Pr(T_i|S_i{\em Rest}_i)$ as follows.
\[\Pr(T_i|S_i{\em Rest}_i) = \Pr(T_i|{\em Rest}_i) \frac{\Pr(T_i|S_{i{\dash}1})}{\Pr(T_i)} \]
We can now use $\frac{\Pr(T_i|S_i)}{\Pr(T_i)}$ as a factor
to modify the tone probability in order to take into account the
silence duration. In Figure~\ref{fig:silences}, we give the factors by
which we adjust the tag probabilities given the amount of silence.
Again, due to sparse of data, we collapse the values of the tone,
editing term and repair tag into six classes: boundary tones, editing term pushes, editing term pops, modification repairs and fresh starts
(without an editing term).
From the figure, we see that if there is no silence between $w_{i-1}$
and $w_i$, the null interpretation for the tone, repair and editing
term tags is preferred. Since the independence assumptions that we
have to make are too strong, we normalize the adjusted tone, editing
term and repair tag probabilities to ensure that they sum to one over
all of the values of the tags.

\section{Example}

To demonstrate how the model works, consider the following example.
\begin{example}{d92a-2.1 utt95}
will take a $\underbrace{\makebox{total of}}_{\makebox[3em][r]{\em reparandum} \hfill}$\ip \et{um let's see} total 
$\underbrace{\makebox{of s-}}_{\makebox[2em][r]{\em reparandum \hfill}}$\ip of 7 hours
\end{example}
The language model considers all possible interpretations (at least
those that do not get pruned) and assigns a probability to each.
Below, we give the probabilities for the correct interpretation of the
word ``um'', given the correct interpretation of the words ``will take a
total of''. For reference, we give a simplified view of the context that is used for each probability.\vspace{0.3em} \\
{\small
$\Pr(T_6$={\bf null}$|$a total of)=0.98 \\
$\Pr(E_6$={\bf Push}$|$a total of)=0.28 \\
$\Pr(R_6$={\bf null}$|$a total of {\bf Push})=1.00 \\
$\Pr(P_6$={\bf UH\_FP}$|$a total of {\bf Push})=0.75 \\
$\Pr(W_6$=um$|$a total of {\bf Push} {\bf UH\_FP})=0.33}
\vspace{0.3em} \\
Given the correct interpretation of the previous words, the
probability of the filled pause ``um'' along with the correct POS tag,
boundary tone tag, and repair tags is 0.0665.

Now lets consider predicting the second instance of ``total'', which is
the first word of the alteration of the first repair, whose editing term ``um let's see'', which ends with a boundary tone, has just finished.
\vspace{0.3em} \\
{\small
$\Pr(T_{10}$={\bf T}$|${\bf Push}\ let's see)=0.93 \\
$\Pr(E_{10}$={\bf Pop}$|${\bf Push} let's see {\bf Tone})=0.79 \\
$\Pr(R_{10}$={\bf M}$|$a total of {\bf Push} let's see {\bf Pop}) = 0.26 \\
$\Pr(O_{10}$=total$|$will take a total of $R_{10}$={\bf Mod})=0.07 \\
$\Pr(L_{10}$=total$|$total of $R_{10}$={\bf Mod})=0.94 \\
$\Pr(C_{10}$={\bf m}$|$will take a $L_{10}$=total/{\bf NN}) = 0.87
4 \\
$\Pr(P_{10}$={\bf NN}$|$will take a $L_{10}$=total/{\bf NN} $C_{10}$={\bf m})=1 \\
$\Pr(W_{10}$=total$|$will take a {\bf NN} $L_{10}$=total $C_{10}$={\bf m})=1}
\vspace{0.3em} \\
Given the correct interpretation of the previous words, the
probability of the word ``total'' along with the correct POS tag,
boundary tone tag, and repair tags is 0.011.

\section{Results}

To demonstrate our model, we use a 6-fold cross validation procedure,
in which we use each sixth of the corpus for testing data, and the
rest for training data.  We start with the word transcriptions of the
Trains corpus, thus allowing us to get a clearer indication of the
performance of our model without having to take into account the poor
performance of speech recognizers on spontaneous speech. All silence
durations are automatically obtained from a word aligner
\cite{Entropic94:aligner}.

Table~\ref{tab:results:pos} shows how POS tagging, discourse marker
identification and perplexity benefit by modeling the speaker's
utterance.
\begin{table}
\setlength{\tabcolsep}{0.2em}
\begin{center}\small
{\small\begin{tabular}{|l|r|r|r|r|} \hline
    &           &                 &\mc{Tones}      \vspace{-0.2em}\\
    &           &\mc{Tones}       &\mc{Repairs}    \vspace{-0.2em}\\
    &\mc{Base}  &\mc{Repairs}     &\mc{Corrections}\vspace{-0.2em}\\
    &\mc{Model} &\mc{Corrections} &\mc{Silences}   \\ \hline \hline
{\em POS Tagging}      &       &       &       \\
\ Error Rate           &  2.95 &  2.86 &  2.69 \\  \hline
{\em Discourse Markers} &       &       &       \\
\ Recall               & 96.60 & 96.60 & 97.14 \\
\ Precision            & 95.76 & 95.86 & 96.31 \\
\ Error Rate           &  7.67 &  7.56 &  6.57 \\ \hline
Perplexity             & 24.35 & 23.05 & 22.45 \\ \hline
\end{tabular}}
\end{center}
\caption{POS Tagging and Perplexity Results}
\label{tab:results:pos}
\end{table}%
The POS tagging results are reported as the percentage of words that
were assigned the wrong tag. The detection of discourse markers is
reported using recall and precision. The recall rate of $X$ is the
number of $X$ events that were correctly determined by the algorithm
over the number of occurrences of $X$. The precision rate is the
number of $X$ events that were correctly determined over the number of
times that the algorithm guessed $X$. The error rate is the number of
$X$ events that the algorithm missed plus the number of $X$ events
that it incorrectly guessed as occurring over the number of $X$
events. The last measure is {\em perplexity}, which is a way of
measuring how well the language model is able to predict the next
word. The perplexity of a test set of $N$ words $w_{1,N}$ is
calculated as follows.
\[ 2^{-\frac1N\sum_{i=1}^N\log_2\Pr(w_i|w_{\rim})} \]

The second column of Table~\ref{tab:results:pos} gives the results of
the POS-based model, the third column gives the results of
incorporating the detection and correction of speech repairs and
detection of intonational phrase boundary tones, and the fourth column
gives the results of adding in silence information. As can be seen,
modeling the user's utterances improves POS tagging, identification of
discourse markers, and word perplexity; with the POS error rate
decreasing by 3.1\% and perplexity by 5.3\%. Furthermore, adding in
silence information to help detect the boundary tones and speech
repairs results in a further improvement, with the overall POS tagging
error rate decreasing by 8.6\% and reducing perplexity by 7.8\%. In
contrast, a word-based trigram backoff model
\cite{Katz87:assp} built with the CMU statistical language modeling
toolkit \cite{Rosenfeld95:arpa} achieved a perplexity of 26.13.
Thus our full language model results in 14.1\% reduction in
perplexity.

Table~\ref{tab:results:ip} gives the results of detecting intonational
boundaries.  
\begin{table}
\setlength{\tabcolsep}{0.2em}
\begin{center}
{\small\begin{tabular}{|l|r|r|r|r|} \hline
  &            &               & \mc{Tones}      \vspace{-0.2em}\\ 
  &            &               & \mc{Repairs}    \vspace{-0.2em}\\ 
  &            & \mc{Tones}    & \mc{Corrections}\vspace{-0.2em}\\ 
  & \mc{Tones} & \mc{Silences} & \mc{Silences}   \\ \hline \hline
{\em Within Turn} &       &       &       \\
Recall            & 64.9  & 70.2  & 70.5  \\
Precision         & 67.4  & 68.7  & 69.4  \\ 
Error Rate        & 66.5  & 61.9  & 60.5  \\ \hline
{\em All Tones}   &       &       &       \\
Recall            & 80.9  & 83.5  & 83.9  \\
Precision         & 81.0  & 81.3  & 81.8  \\ 
Error Rate        & 38.0  & 35.7  & 34.8  \\ \hline
Perplexity        & 24.12 & 23.78 & 22.45 \\ \hline
\end{tabular}}
\end{center}
\caption{\label{tab:results:ip}Detecting Intonational Phrases}
\end{table}
The second column gives the results of adding the boundary tone
detection to the POS model, the third column adds silence
information, and the fourth column adds speech repair detection and
correction. We see that adding in silence information gives a
noticeable improvement in detecting boundary tones.  Furthermore,
adding in the speech repair detection and correction further improves
the results of identifying boundary tones.  Hence to detect
intonational phrase boundaries in spontaneous speech, one should also
model speech repairs.

Table~\ref{tab:detection} gives the results of detecting and
correcting speech repairs. The detection results report the number of
repairs that were detected, regardless of whether the type of repair
(e.g.~modification repair versus abridged repair) was properly
determined.
\begin{table}
\setlength{\tabcolsep}{0.2em}
\begin{center}
{\small\begin{tabular}{|l|r|r|r|r|} \hline
&             &             &                &\mc{Tones}      \vspace{-0.2em}\\
&             &             &\mc{Repairs}    &\mc{Repairs}    \vspace{-0.2em}\\
&             &\mc{Repairs} &\mc{Corrections}&\mc{Corrections}\vspace{-0.2em}\\
&\mc{Repairs} &\mc{Silences}&\mc{Silences}   &\mc{Silences}  \\ \hline \hline
{\em Detection}           &       &       &       &       \\
\ Recall                  &  67.9 &  72.7 &  75.7 &  77.0 \\
\ Precision               &  80.6 &  77.9 &  80.8 &  84.8 \\
\ Error Rate              &  48.5 &  47.9 &  42.4 &  36.8 \\ \hline
{\em Correction}          &       &       &       &       \\
\ Recall                  &       &       &  62.4 &  65.0 \\
\ Precision               &       &       &  66.6 &  71.5 \\
\ Error Rate              &       &       &  68.9 &  60.9 \\ \hline
Perplexity                & 24.11 & 23.72 & 23.04 & 22.45 \\ \hline
\end{tabular}}
\end{center}
\caption{\label{tab:detection}Detecting and Correcting Speech Repairs}
\end{table}
The second column gives the results of adding speech repair detection
to the POS model. The third column adds in silence information. Unlike
the case for boundary tones, adding silence does not have much of an
effect.\footnote{Silence has a bigger effect on detection and
correction if boundary tones are modeled.} The fourth column adds in
speech repair correction, and shows that taking into account the
correction, gives better detection rates \cite
{HeemanLokenkimAllen96:icslp}. The fifth column adds in boundary tone
detection, which improves both the detection and correction of speech
repairs.

\section{Comparison to Other Work}

Comparing the performance of this model to others that have been
proposed in the literature is very difficult, due to differences in
corpora, and different input assumptions.  However, it is useful to
compare the different techniques that are used.

Bear~\etal~\shortcite{Bear-etal92:acl} used a simple pattern matching
approach on ATIS word transcriptions.  They exclude all turns that
have a repair that just consists of a filled pause or word fragment.
On this subset they obtained a correction recall rate of 43\% and a
precision of 50\%.

Nakatani and Hirschberg \shortcite{NakataniHirschberg94:jasa} examined
how speech repairs can be detected using a variety of information,
including acoustic, presence of word matchings, and POS tags.  Using
these clues they were able to train a decision tree which achieved a
recall rate of 86.1\% and a precision of 92.1\% on a set of turns in
which each turn contained at least one speech repair.

Stolcke and Shriberg \shortcite{StolckeShriberg96:icassp} examined
whether perplexity can be improved by modeling simple types of speech
repairs in a language model. They find that doing so actually makes
perplexity worse, and they attribute this to not having a linguistic
segmentation available, which would help in modeling filled pauses. We
feel that speech repair modeling must be combined with detecting
utterance boundaries and discourse markers, and should take advantage
of acoustic information.

For detecting boundary tones, the model of Wightman and Ostendorf
\shortcite {WightmanOstendorf94:ieee} achieves a recall rate of 78.1\%
and a precision of 76.8\%.  Their better performance is partly
attributed to richer (speaker dependent) acoustic modeling,
including phoneme duration, energy, and pitch.  However, their model
was trained and tested on professionally read speech, rather than
spontaneous speech.

Wang and Hirschberg~\shortcite{WangHirschberg92:csl} did employ
spontaneous speech, namely, the ATIS corpus. For turn-internal
boundary tones, they achieved a recall rate of 38.5\% and a precision of
72.9\% using a decision tree approach that combined both textual
features, such as POS tags, and syntactic constituents with
intonational features. One explanation for the difference in
performance was that our model was trained on approximately ten times
as much data. Secondly, their decision trees are used to classify each
data point independently of the next, whereas we find the best
interpretation over the entire turn, and incorporate speech repairs.

The models of Kompe~\etal~\shortcite{Kompe-etal94:icassp} and
Mast~\etal~\shortcite{Mast-etal96:icslp} are the most similar to our
model in terms of incorporating a language model. Mast~\etal~achieve a
recall rate of 85.0\% and a precision of 53.1\% on identifying dialog
acts in a German corpus. Their model employs richer acoustic modeling,
however, it does not account for other aspects of utterance modeling,
such as speech repairs.

\section{Conclusion}

In this paper, we have shown that the problems of identifying
intonational boundaries and discourse markers, and resolving speech
repairs can be tackled by a statistical language model, which uses
local context. We have also shown that these tasks, along with POS
tagging, should be resolved together. Since our model can give a
probability estimate for the next word, it can be used as the language
model for a speech recognizer. In terms of perplexity, our model gives
a 14\% improvement over word-based language models. Part of this
improvement is due to being able to exploit silence durations, which
traditional word-based language models tend to ignore. Our next step
is to incorporate this model into a speech recognizer in order to
validate that the improved perplexity does in fact lead to a better
word recognition rate.

\section{Acknowledgments}

This material is based upon work supported by the NSF under grant
IRI-9623665 and by ONR under grant N00014-95-1-1088. Final preparation
of this paper was done while the first author was visiting CNET,
France T\'el\'ecom.



\def\thebibliography#1{\section*{References}\list
 {}{\setlength{\labelwidth}{0pt}\setlength{\leftmargin}{\parindent}
 \setlength{\itemindent}{-\parindent}}
 \itemsep=-2pt
 \def\baselinestretch{0.95}\small
 \vskip -9pt
 \advance\leftmargin\labelsep
 \def\newblock{\hskip .11em plus .33em minus -.07em}
 \sloppy\clubpenalty4000\widowpenalty4000
 \sfcode`\.=1000\relax}

\end{document}